\documentclass[sn-mathphys,Numbered]{sn-jnl}

\usepackage{graphicx}%
\usepackage{multirow}%
\usepackage{amsmath,amssymb,amsfonts}%
\usepackage{amsthm}%
\usepackage{mathrsfs}%
\usepackage[title]{appendix}%
\usepackage{xcolor}%
\usepackage{textcomp}%
\usepackage{manyfoot}%
\usepackage{booktabs}%
\usepackage{algorithm}%
\usepackage{algorithmicx}%
\usepackage{algpseudocode}%
\usepackage{listings}%

\usepackage{amsmath,graphicx}
\usepackage{ulem,color,multirow,hhline,url}

\def\eg{\textit{e.g.~}}

\def\ie{\textit{i.e.~}}
\def\etal{\textit{et al.~}}

\begin{document}

\title[Removing cloud shadows from ground-based solar imagery]{Removing cloud shadows from ground-based solar imagery}


\author[1]{\fnm{Amal} \sur{Chaoui}}\email{amal.chaoui@centrale.centralelille.fr}

\author[2]{\fnm{Jay Paul} \sur{Morgan}}\email{jay.morgan@univ-tln.fr}

\author*[2]{\fnm{Adeline} \sur{Paiement}}\email{adeline.paiement@univ-tln.fr}

\author[3]{\fnm{Jean} \sur{Aboudarham}}\email{Jean.Aboudarham@obspm.fr}

\affil[1]{\orgname{Université de Lille, Centrale Lille}, \orgaddress{\city{Lille}, \country{France}}}

\affil[2]{\orgname{Université de Toulon, Aix Marseille Univ, CNRS, LIS}, \orgaddress{\city{Marseille}, \country{France}}}

\affil[3]{\orgname{Observatoire de Paris/PSL, LESIA, CNRS}, \orgaddress{\city{Paris}, \country{France}}}


\abstract{The study and prediction of space weather entails the analysis of solar images showing structures of the Sun's atmosphere. When imaged from the Earth's ground, images may be polluted by terrestrial clouds which hinder the detection of solar structures. We propose a new method to remove cloud shadows, based on a U-Net architecture, and compare classical supervision with conditional GAN. We evaluate our method on two different imaging modalities, using both real images and a new dataset of synthetic clouds. Quantitative assessments are obtained through image quality indices (RMSE, PSNR, SSIM, and FID).
We demonstrate improved results with regards to the traditional cloud removal technique and a sparse coding baseline, on different cloud types and textures.}




\maketitle

\section*{Acknowledgments}

This work was funded by ANR grant N° ANR-20-CE23-0014-01.
One of the authors (JA) thanks CNES (Centre National d'Etudes Spatiale, the French space agency) for its support in this project.


\section{Introduction and previous works}
\label{sec:intro}

\begin{figure}
\begin{center}
    \includegraphics[width=0.9\linewidth]{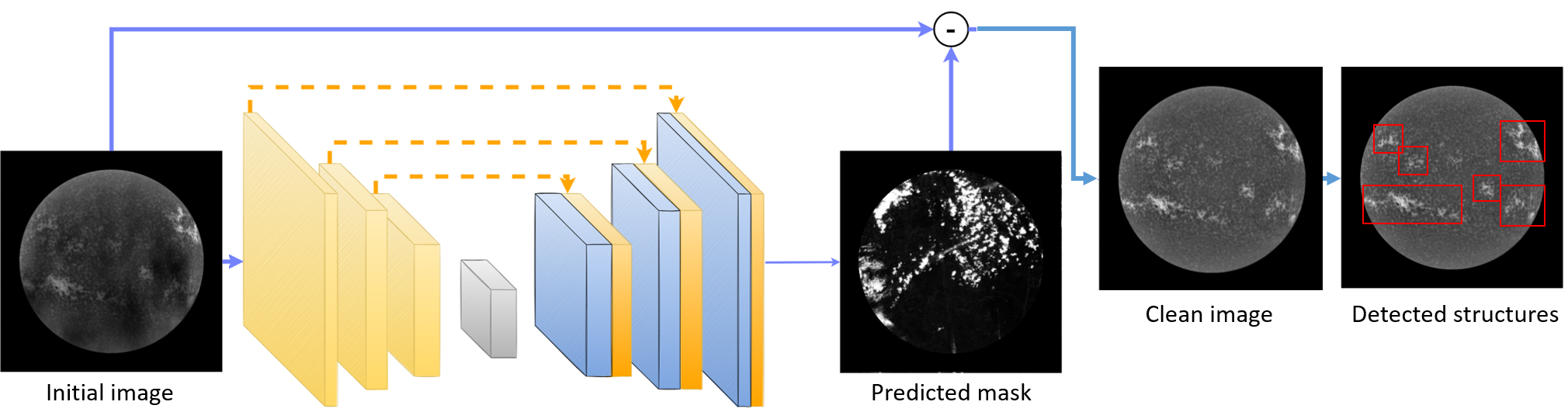}
\end{center}
   \caption{Overview of cloud removal from solar imagery. A cloud mask is predicted and subtracted from the original image, to prepare for a subsequent detection of solar structures (here ARs in red boxes). Our proposed method only deals with the cloud removal stage.}
\label{fig:overview}
\end{figure}

This paper addresses the problem of cleaning ground-based solar imagery from cloud shadow contaminants, as illustrated in Fig.~\ref{fig:overview}, in support to forthcoming analyses with solar physics and space weather applications.
Monitoring the activity of the Sun is a task of growing importance in the frame of space weather research. The main sources of severe space weather disturbances at Earth are energetic outbursts from the Sun, \eg solar flares and coronal mass ejections. They are strongly associated with solar structures of the chromosphere (\eg active regions (AR) and filaments, as illustrated in Fig.~\ref{fig:vignettes}), visible from the Earth ground in the light emitted by ionised calcium (Ca-II, at a wavelength of 3933.7{\AA}) 
and the H-$\alpha$ line of neutral hydrogen (6562.8{\AA}). Despite their affordability, images taken from the ground present a particular difficulty: they are prone to be tainted by cloud shadows when the sky is not clear. These shadows are rarely uniform and may be mistaken for solar structures (\eg filaments). Thus, it is substantial to remove them before automatic solar structure detection and monitoring.

Very few previous works addressed this problem. \cite{feng2014,fuller2004} designed denoising methods for H-$\alpha$ images.
Feng \etal \cite{feng2014} proposed a metric for assessing the level of cloud coverage in an image by considering the homogeneity of image intensity between different quadrants of the solar disk. Additionally, they proposed a denoising method that first estimates the pixel-wise transmittance of clouds $\hat{I}_t$ from a cloud-free temporal neighbour using a low-pass morphological filter to smooth out solar structures (\ie filaments). Second, it uses it to normalise the original image $I_\text{ini}$:
\begin{equation}
    \hat{I}_\text{clean} = I_\text{ini} / \hat{I}_t
    \label{eq:transmitance}
\end{equation}
In \cite{fuller2004}, Fuller and Aboudarham perform two stages of median filtering to identify large scale intensity variations that correspond to clouds, while thresholding out the solar structures.
In both methods, the use of low-pass filtering to distinguish clouds from solar structures is a possible limitation when their spatial scales are similar in at least one direction.

In other application domains, some works removed large contaminants, such as clouds and shadows, from remote sensing or natural images. Due to clouds transparency, the task of removing clouds in Sun observations resembles to a degree to these tasks. While there is a large literature on denoising methods \eg \cite{chen2020oct}, we focus on the removal of large and semi-transparent contaminants, as this is the closest to our task. Methods that perform inpainting over opaque contaminants, such as \cite{zhengGAN} for thick clouds, are not considered to avoid introducing artefacts in the study of solar structures (images with thick clouds are therefore not usable). We also discard methods that exploit the physics of image formation for natural images, \eg \cite{Finlayson2006,Yang2012,Guo2013,Yu2017}, as solar imaging follows a different physics. Thin cloud and shadow removal methods are good candidates for our task of removing semi-transparent clouds while preserving the background solar structures.

Some shadow suppression methods \eg \cite{sasishadowsparse} rely upon the sparse approximation theory. A local dictionary of atoms is learnt from image patches as a basis for sparse coding. This allows decomposing the image into reflectance (invariant to shading), geometry (fine structural details), and shadow components, the latter being excluded during image reconstruction.
The use of local dictionaries, rather than generic ones (\eg wavelets, curvelets), helps in learning atoms that adapt to each single image. However, this approach cannot gather further insights from multiple examples.

This limitation is addressed by (deep) learning-based methods. \cite{fan2019cnn,deshadow_net} use encoder-decoder deep neural networks (DNN) to remove shadows by learning a mapping between the initial images $I_{ini}$ and their binary mask $\hat{M}$ of shadow ratio, which is reminiscent of the transmittance image of \cite{feng2014}:
\begin{equation}
    \hat{I}_\text{clean} = I_\text{ini} / \hat{M}
    \label{eq:shadow_mask}
\end{equation}
\cite{st_gan} and pix2pix \cite{pix2pixPaper} in \cite{zhengGAN} rely on skip connections (U-Net-style) to preserve spatial resolution, and a conditional generative adversarial network (C-GAN) for more efficient training. They obtain state-of-the-art (SoTA) results in shadow and thin cloud removal. In \cite{st_gan}, an additional C-GAN outputs the denoised image from predicted shadow mask, while pix2pix of \cite{zhengGAN} outputs the cleaned image directly.


We explore the applicability of these denoising methods for the new application of removing cloud shadows from solar imagery. We evaluate sparse representation-based and deep learning-based approaches, and compare against \cite{feng2014,fuller2004} (Section \ref{sec:results}). Inspired by \cite{st_gan,zhengGAN}, we use a combination of U-Net and C-GAN (Section \ref{sec:method}) to obtain the new SoTA for this task.

The methods are comparatively assessed on two imaging modalities. 
To support these experiments, we introduce two new datasets of Sun images with artificial and real clouds.

\section{Data}
\label{sec:data}

Our images are from Paris-Meudon (PM) observatory\footnote{The Paris-Meudon observatory is part of Observatoire de Paris-PSL: \url{https://www.observatoiredeparis.psl.eu/-observatoire-de-paris-.html?lang=en}}. They are $1024\times1024$~px full-disk spectroheliograms. Care has been taken in their collection to ensure a good representativity. We used images of two modalities: Ca-II (for studying ARs), and H-$\alpha$ (for filaments), as highlighted in 
Fig.~\ref{fig:vignettes}. These are major modalities in ground-based solar imaging. Observations span different periods of the solar cycle in order to reflect variations in the Sun's activity and number / size of the solar structures that may be observed at the surface of the Sun. 96 Ca-II and 111 H-$\alpha$ images are sampled between May 2002 and June 2017. Sampling uses the metric of cloud coverage level proposed by \cite{feng2014} to select the images that are tainted with cloud shadows.

PM images are taken in principle during clear sky days. However, even during those days, small thin clouds emerging in the horizon cannot be prevented and result in light cloud stains. Cloud shadows can come with different textures.
In PM observatory, the Sun disk is scanned by a thin moving slit over 20 to 100 sec.~\cite{meudonSpectroheliograph}. In case of fast enough cloud motion, due to the slit movement alongside the motion of clouds, cloud spots may be shifted along the scanning process, creating straight lines on the image (denoted as ``streaked'' texture). Slower cloud motions result in the ``fluffy'' texture that is customary of many terrestrial clouds. It is also a common texture in solar observations from observatories that use 2D cameras rather than a spectrograph. The ``streaked'' texture may provide more challenges to denoising methods, due to having different spatial scales in different directions. Fig.~\ref{fig:vignettes} illustrates samples of streaked and fluffy textures with different cloud cover thicknesses and streak orientations.

\begin{figure}
\begin{center}
    \includegraphics[width=0.9\linewidth]{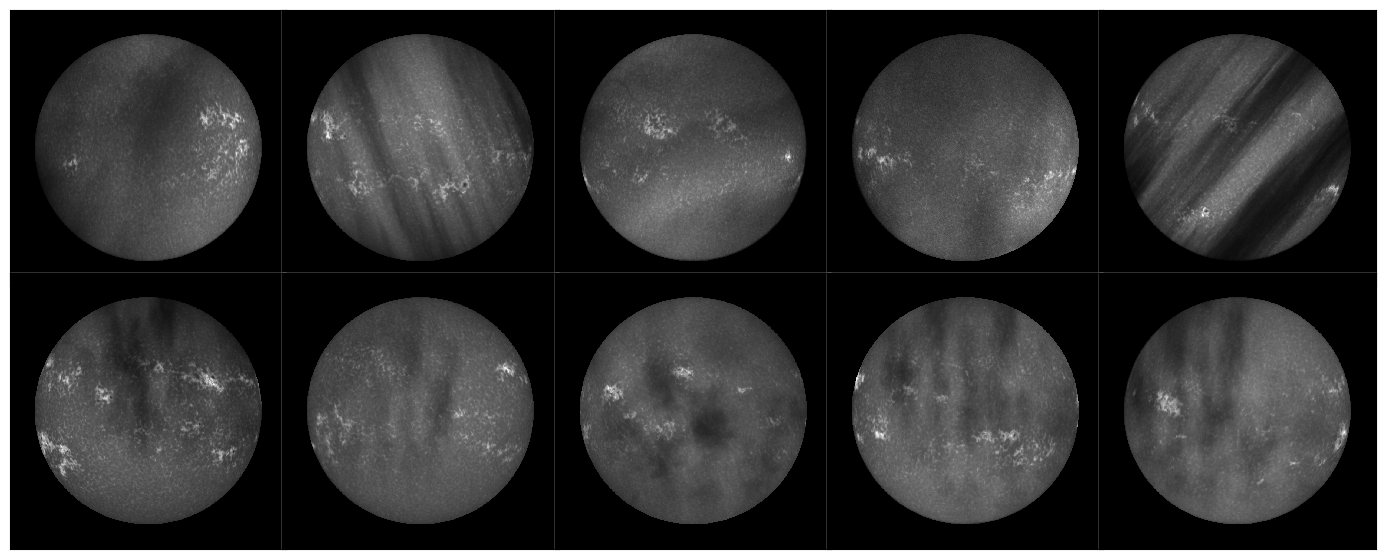}
\end{center}
   \caption{Examples of real (top) and synthetic (bottom) cloud contamination in Ca-II images.} 
\label{fig:clouds_examples}
\end{figure}

The cloudy images don't have cloud-free counterparts for training DNNs. Thus, we introduce a dataset of real (initially cloud-free) images with added synthetic cloud shadows. Using the cloud coverage level metric of \cite{feng2014}, we selected reasonably cloud-free images in the same time period as above. 
A basic cloud texture was obtained from \cite{cloudtexture}. For each cloud-free image, a new cloud shadow was composed by summing a random number (in range [2, 3]) of duplicates of the basic texture with horizontal and vertical random resizing between 50-100\% of the original size, random flips (around one or two axes), and random transparencies (in range [0.1, 0.4]). It was then overlapped to the cloud-free image. This procedure provides realistic images of fluffy and semi-transparent cloud shadow contamination. They are diverse and representative of possible cloud shapes and opacity levels.
The Ca-II and H-$\alpha$ synthetic datasets comprise respectively 319 and 367 pairs of shadow/shadow-free images, split into 179/44/96 and 205/51/111 training/validation/testing pairs.
The combination of our real and artificial cloud datasets provides a variety of streaked and fluffy cloud textures. They are publicly available\footnote{\url{https://doi.org/10.5281/zenodo.7684200}}, along with the pretrained cloud-removal DNNs\footnote{The source code and model weights are available via \url{https://gitlab.lis-lab.fr/presage/cloud-removal-from-solar-imagery.}}.

Images have their intensity normalised in range $[0, 1]$, centred solar disks with normalised radii, and background pixels are set to zero outside of the solar disk to remove any distracting features from background stars and imaging artefacts. 

\section{Methodology}
\label{sec:method}

We adopt the deep learning approach due to its SoTA results in large contaminants removal. As in \cite{st_gan,pix2pixPaper}, we use skip connections to preserve spatial resolution and details\footnote{\label{foot:unet}We use the U-Net implementation from \url{https://github.com/mateuszbuda/brain-segmentation-pytorch}.}. Inspired by the results of previous methods, we compare the classical fully supervised and the C-GAN training strategies.

We experiment with our DNNs predicting either the cleaned image directly, or a cloud shadow mask, to be used as in \cite{feng2014,fan2019cnn,deshadow_net}. To avoid division by zero and improve stability, we compute the clean image as:
\begin{equation}
    \hat{I}_\text{clean} = \frac{I_\text{ini}}{\hat{M} + \epsilon}
    \label{eq:division}
\end{equation}
We empirically verified that the exact value of $\epsilon$ has little impact on the results, and we set it to $1e^{-5}$ in our experiments.
Regressing a shadow mask instead of the cleaned image directly is motivated by the physics of solar imaging, as in \cite{feng2014}. In addition, estimating a transmittance or shadow ratio, rather than the pixel's absolute intensity, reduces the risk to create artefacts that would hinder the forthcoming study of solar structures. Finally, estimating a shadow mask should be a less complex task for the DNN, which may be learnt more efficiently. Indeed, reproducing the exact patterns of the more complex and diverse solar structures in the output image would be an unnecessary additional task for the DNN.

With Eq.~\ref{eq:division}, the loss gradient varies as $I_\text{ini}/\hat{M}^2$, meaning that the darkest pixels of the images, where cloud thickness is high, see higher loss gradients and require a finer adjustment of their mask value. In an attempt to improve stability during training through more consistent loss gradient magnitudes, we also experiment with an alternative implementation, \ie adding in the intensity that was removed by clouds. This is similar to predicting a residual mask:
\begin{equation}
    \hat{I}_\text{clean} = I_\text{ini} + \hat{M}
    \label{eq:subtraction}
\end{equation}

We compare two training strategies: a fully supervised DNN (denoted as FS) and a C-GAN following the pix2pix setup. For ease of comparison, the FS DNN and C-GAN generator have the same U-Net-style architecture$^{\ref{foot:unet}}$: down-sampling and up-sampling blocks along with skip connections, each block as defined in \cite{Ronneberger2015}.
We experiment with a varying number of blocks to determine the optimal complexity and receptive field needed to account for both large scale contaminants and smaller scale solar structures. Due to memory constraints, we go up to 6 down/up-sampling blocks, and find it to be the best performing architecture (see supplementary materials). In the future, and on a different hardware, more blocks could be tried.

We use a linear activation in the last block when predicting $\hat{I}_\text{clean}$ directly or $\hat{M}$ as in Eq.~\ref{eq:subtraction}. For Eq.~\ref{eq:division}, we find that imposing the output to be in range $[0,1]$ is necessary to avoid instability issues, so we use $0.5 \times \tanh + 0.5$ as activation.

The discriminator in the C-GAN is the same as in pix2pix \cite{pix2pixPaper}: a PatchGAN network that considers image patches (small receptive fields) to capture high-frequencies. We experimentally verified that the recommended receptive field of $70\times70$ is also optimal in our application. The discriminator is applied to the cleaned image rather than the predicted shadow mask to assess the preservation of solar structures.


For the C-GAN, we use the loss function suggested in \cite{pix2pixPaper}.
The loss function of the FS network and of the C-GAN generator uses the pixel-wise $L_1$-norm since it promotes sparsity and encourages less blurring as noted in \cite{pix2pixPaper}. The pixel-level loss is computed between the target $I_t$ and cleaned image $\hat{I}_\text{clean}$:
\begin{equation}
    \mathcal{L}_{L_1}(I_t, \hat{I}_\text{clean}) = \frac{1}{N} \sum_{i=1}^N |I_t^i - \hat{I}_\text{clean}^i|
    \label{eq:dnnU-NetLoss}
\end{equation}

For both DNNs, we use the default U-Net$^{\ref{foot:unet}}$ training scheme and batch size of 3. 
We train for 100 epochs and test on the epoch with the best validation loss (FS DNN) or generator's training loss (C-GAN, following the usual practice for GANs). 
During testing of the C-GAN, the generator is run with dropout, and batch normalisation considers the statistics of the batch, as in \cite{pix2pixPaper}.

\section{Results}
\label{sec:results}

\subsection{Choice of DNN output}

\begin{figure}[t]
    \centering
    \includegraphics[width=\linewidth]{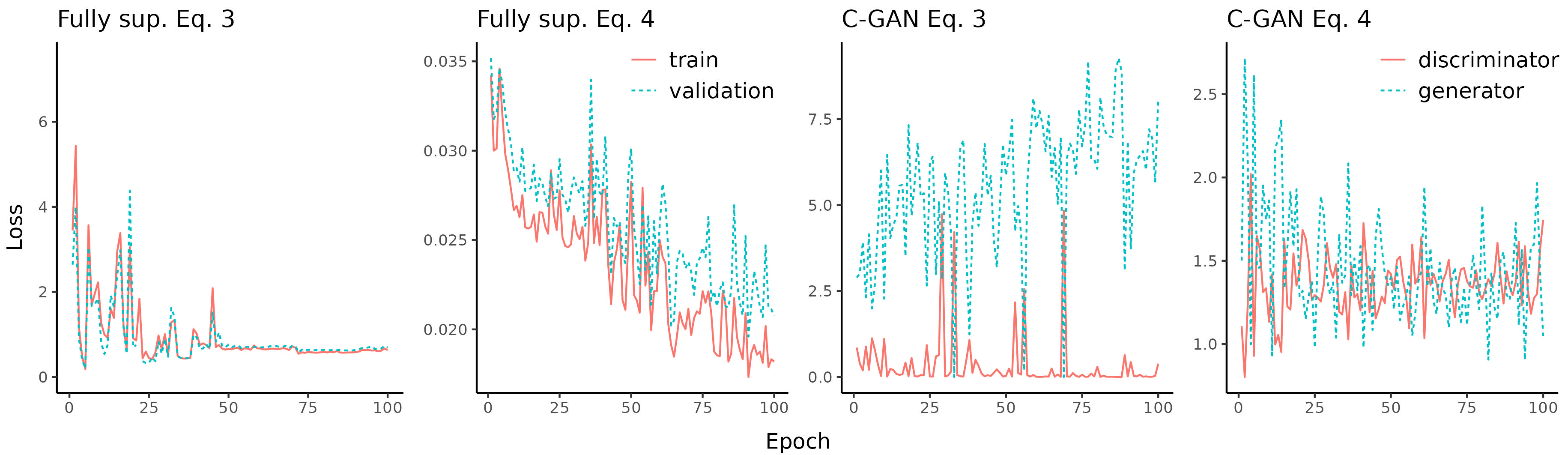}
   \caption{Loss curves when training the proposed models using Eqs.~\ref{eq:division} and \ref{eq:subtraction}.}
\label{fig:comp_div_sub}
\end{figure}

\begin{figure}[t]
    \centering
    \includegraphics[width=1\textwidth]{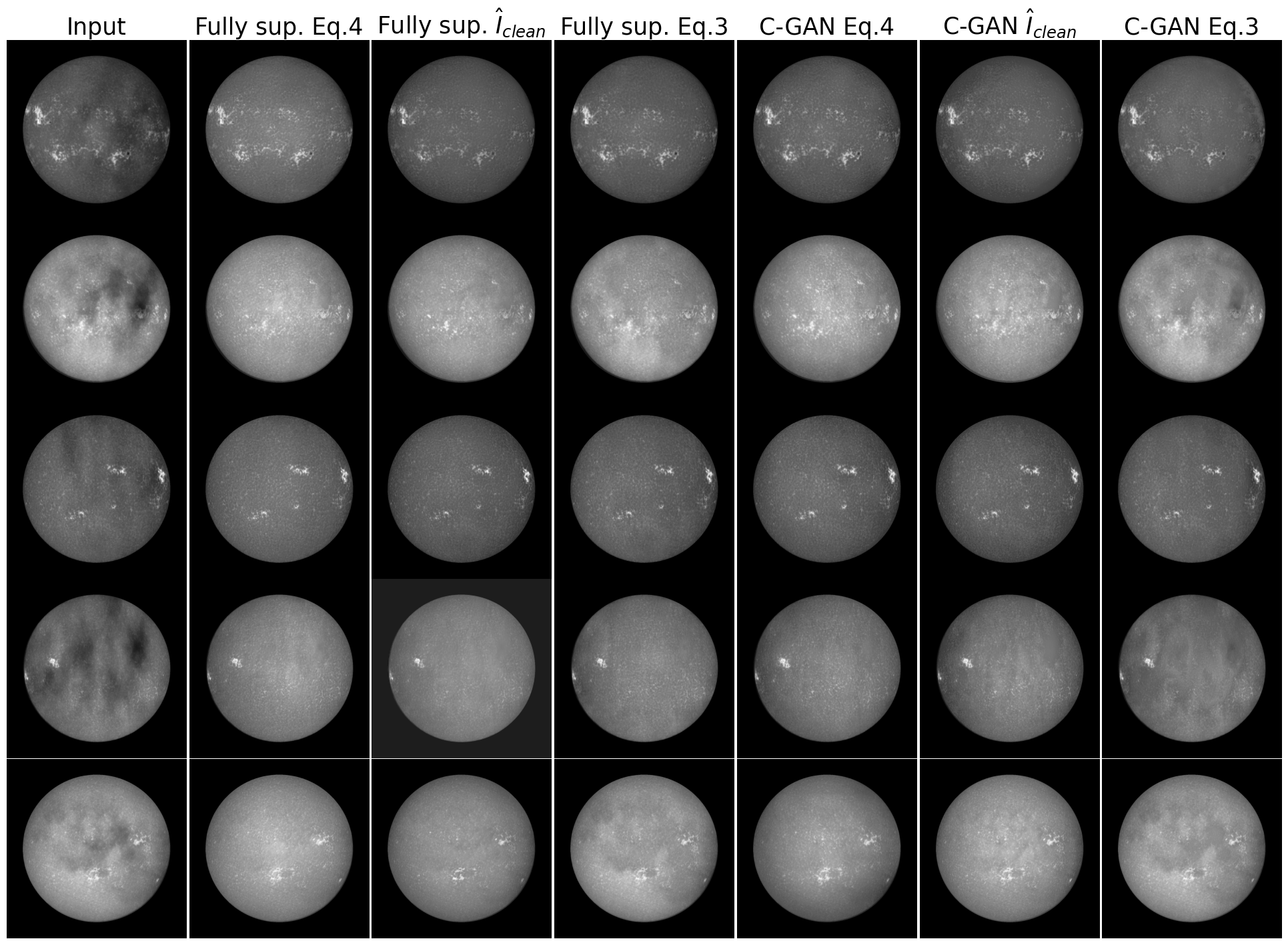}
    \caption{Comparative illustration of the three DNN outputs on some Ca-II images with synthetic clouds. We display the results of a random run of each network, since all runs produced roughly the same results.}
    \label{fig:comp_outputs}
\end{figure}

We compare the three possible outputs of our DNNs, \ie direct cleaned image and shadow masks to be used as in Eqs.~\ref{eq:division} (transmittance) and \ref{eq:subtraction} (residual). Table \ref{tab:outputs} reports the metrics on the Ca-II synthetic dataset and Fig.~\ref{fig:comp_outputs} provides some illustration examples. 
For both DNNs, Eq.~\ref{eq:division} is less stable than Eq.~\ref{eq:subtraction} during training, with poorer metrics and more noisy loss curves as shown in Fig.~\ref{fig:comp_div_sub}. 
This more unstable training results in a poor convergence. Poorer results are observed in the sensitive cases of strong cloud coverage (\ie low values in $I_\text{ini}$) where more subtle adjustments are needed in the mask's values to correct for the clouds. 
Predicting a residual mask $\hat{M}$ for Eq.~\ref{eq:subtraction} performs slightly better than the direct prediction of $\hat{I}_\text{clean}$, which, in rare cases, created artefacts that may be confused with solar structures. The better performance of Eq.~\ref{eq:subtraction} is in line with other works on residual prediction.
Following these observations, we choose Eq.~\ref{eq:subtraction} to be used in the rest of this paper.

\begin{table}[t]
   \caption{Comparison of the three DNN outputs on the Ca-II synthesised dataset. Results are in format mean(std) over 10 runs. SSIM and RMSE are as $e-2$.}
   \begin{tabular}{cccccc}
       \hline
       DNN & Output & PSNR $\uparrow$ & SSIM $\uparrow$ & RMSE $\downarrow$ & FID $\downarrow$ \\
       \hline
       \multirow{3}{1.6cm}{Fully sup.} & $\hat{I}_\text{clean}$ & 30.2(0.4) & 98.2(0.5) & 3.9(0.1) & 17.2(5.0) \\
       & $\hat{M}$ Eq.~\ref{eq:division} & 28.9(0.2) & 98.8(0.1) & 4.5(0.1) & 20.2(2.4) \\
       & $\hat{M}$ Eq.~\ref{eq:subtraction} & \bf{30.6(0.3)} & \bf{98.9(0.1)} & \bf{3.7(0.1)} & \bf{16.8(5.3)} \\
       \hline
       \multirow{3}{1.6cm}{C-GAN} & $\hat{I}_\text{clean}$ &29.4(0.6) & 97.7(0.4) & 4.1(0.2) & 22.0(10.2) \\
       & $\hat{M}$ Eq.~\ref{eq:division} & 27.3(0.6) & 97.6(0.8) & 5.1(0.2) & 49.0(10.8) \\
       & $\hat{M}$ Eq.~\ref{eq:subtraction} & \bf{30.0(0.4)} & \bf{98.8(0.2)} & \bf{3.9(0.1)} & \bf{18.4(6.3)} \\
       \hline
   \end{tabular}
   \label{tab:outputs}
\end{table}

\subsection{Comparative evaluation of the methods}

We compare the proposed DNNs against the SoTA \cite{feng2014,fuller2004} and the sparse approximation method of \cite{sasishadowsparse} as non-learning baselines. \cite{feng2014} requires a clean temporal neighbour image (within 24 hours), which may not be always available. We consider these cases a failure and report their numbers. The first reflectance step of \cite{sasishadowsparse} is skipped, as reflection plays no role in solar imaging. We use the Peak Signal-to-Noise Ratio (PSNR), Structured Similarity Index Measure (SSIM) and Root Mean Squared Error (RMSE) metrics to assess the quality of the predicted shadow-free images. In addition, the Fréchet Inception Distance (FID) metric \cite{heusel2017gans} is used to compare the distribution of cleaned images against that of targets. 
For \cite{feng2014}, the metrics are computed over non-failure cases. For the DNNs, they are averaged over 10 runs.

\begin{table}[t]
   \caption{Performance on synthesised datasets. Results are in format mean(std) over 10 runs for the DNNs. SSIM and RMSE are as $e-2$.}
   \begin{tabular}{ccccccc}
       \hline
       Data & \multicolumn{2}{c}{Method} & PSNR $\uparrow$ & SSIM $\uparrow$ & RMSE $\downarrow$ & FID $\downarrow$ \\
       \hline
       \multirow{5}{*}{Ca-II} & \multicolumn{2}{c}{\cite{feng2014} (15 failures)} & 21.90 & 94.8 & 8.6 & 104.8 \\
       & \multicolumn{2}{c}{\cite{fuller2004}} & 26.0 & 98.1 & 5.4 & 70.3 \\
       & \multicolumn{2}{c}{\cite{sasishadowsparse}} & 23.2 & 92.7 & 7.3 & 45.5 \\
       \hhline{~--~~~}
           & \multicolumn{2}{c}{Fully sup.} & \bf{30.6(0.3)} & \bf{98.9(0.1)} & \bf{3.7(0.1)} & \bf{16.8(5.3)} \\
       & \multicolumn{2}{c}{C-GAN} & 30.0(0.4) & 98.8(0.2) & 3.9(0.1) & 18.4(6.3) \\
       \hline
       \multirow{5}{*}{H-$\alpha$} & \multicolumn{2}{c}{\cite{feng2014} (5 failures)} & 14.8 & 91.4 & 19.6 & 118.3 \\
       & \multicolumn{2}{c}{\cite{fuller2004}} & 23.3 & 98.4 & 7.0 & 134.7 \\
       & \multicolumn{2}{c}{\cite{sasishadowsparse}} & 21.0 & 96.0 & 9.2 & 110.9 \\
       \hhline{~--~~~}
       & \multicolumn{2}{c}{Fully sup.} & \bf{28.6(0.3)} & \bf{98.9(0.1)} & \bf{4.5(0.2)} & 33.0(4.9) \\
       & \multicolumn{2}{c}{C-GAN} & 28.3(0.5) & 98.9(0.2) & 4.6(0.2) & \bf{37.5(3.5)} \\
       \hline
   \end{tabular}
   \label{tab:comparative_results}
\end{table}

\begin{figure}[t]
    \centering
    \includegraphics[width=0.9\linewidth]{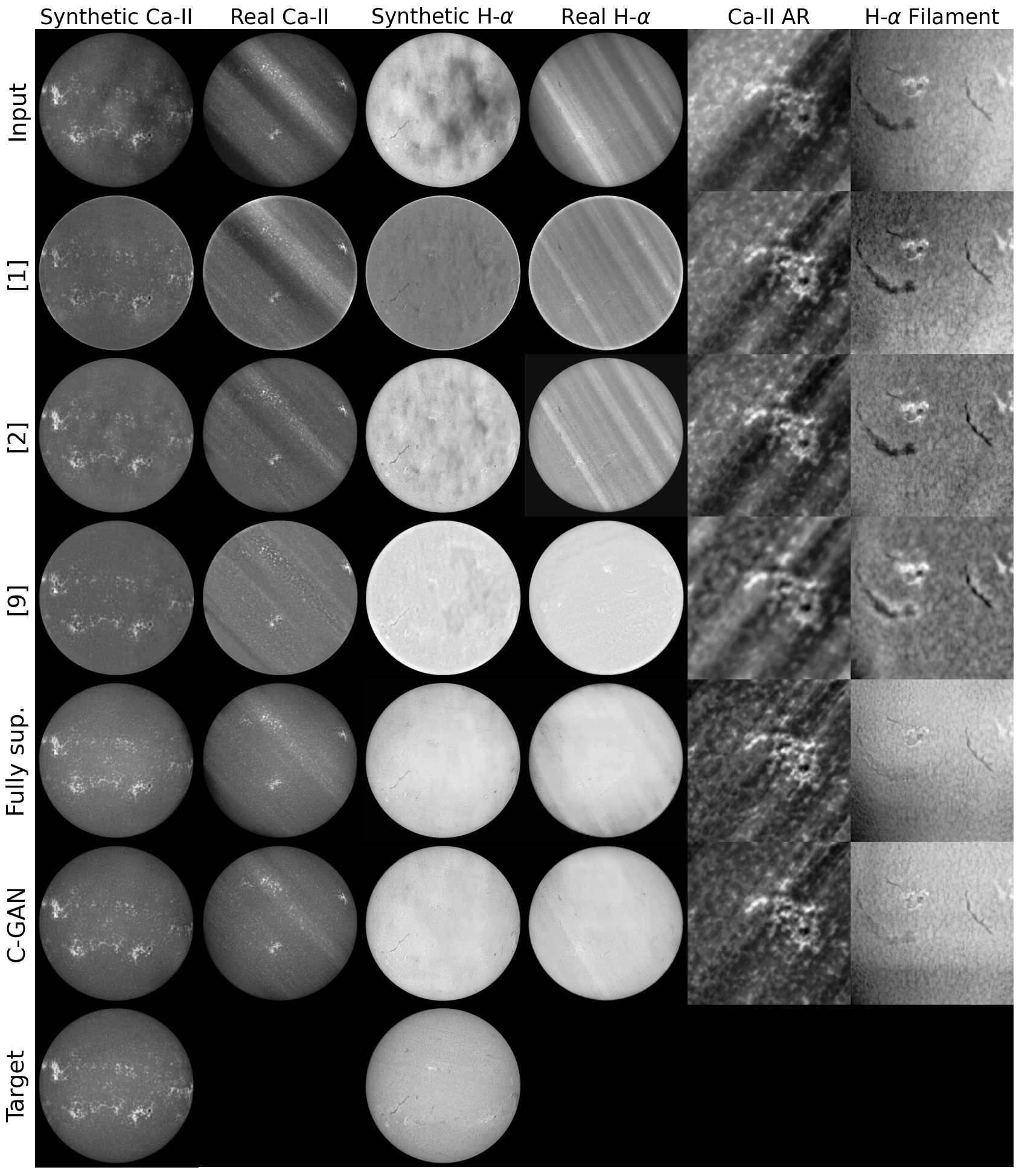}
   \caption{Examples of removal of real and synthetic clouds from Ca-II and H-$\alpha$ images by compared algorithms. Images with real clouds (Real Ca-II, Real H-$\alpha$, Ca-II AR, H-$\alpha$ Filament) do not have a clean target counterpart, as explained in Section \ref{sec:data}. Therefore, for these columns, the target row is left blank.}
\label{fig:vignettes}
\end{figure}

Table \ref{tab:comparative_results} presents results on the synthetic cloud datasets, 
and Fig.~\ref{fig:vignettes} illustrates how the algorithms perform on real/synthetic clouds. More illustrations are provided in the supplementary materials. 
\cite{feng2014} obtains the worst results on both imaging modalities. It struggles to preserve solar structures when their spatial scale is similar to the scale of clouds in at least one direction, as often happens in the case of streaked textures. It is also less robust to strong cloud thickness. \cite{fuller2004} also struggles with the spatial scale of streaked clouds. 
\cite{sasishadowsparse} fails to recover the underlying solar structures and results in poor contrast. 

The proposed DNNs outperform non-learning methods on all datasets. On synthesised images, the DNN's respective metrics are not significantly different. Upon visual inspection of artificial and real images, FS produces a slightly more uniform contrast across the solar disk, while the C-GAN's discriminator, focusing on small patches, allows more diversity of contrasts. However, FS tends to over-compensate strong clouds resulting in a lower contrast there, while C-GAN benefits from enforcing a patch-wise visually pleasant output.

It is worth noting that in our real-cloud dataset, most clouds are streaked, while the DNNs are trained on the fluffy synthetic textures. Both DNNs generalised well to this new cloud texture. While, in some cases, faint streaks remain from the cleaned images, this happens for the strongest clouds and generally doesn't prevent the restoration of underlying solar structures. C-GAN performs slightly better than FS in this particular case of strong streaked clouds. We expect that in the future, training the DNNs with examples of streaked clouds may help in handling the strongest streaked clouds.

Both DNNs provide a detailed restoration of ARs, with the aforementioned contrast differences. 
Filaments suffer from a lower contrast, likely due to their dark appearance, making them harder to distinguish from clouds. This is confirmed by the thicker and therefore blurrier ones being the most affected (erased) by this contrast reduction. C-GAN preserves slightly better these thick filaments.



\section{Conclusion}
\label{sec:conclusion}

We investigated deep learning methods for removing large cloud shadow contaminants from ground-based solar imagery. We adopted the U-Net-style architecture for more detailed outputs. We compared three possible inference methods (\ie direct and through a transmittance or residual mask), and two training setups: fully-supervised and C-GAN. 
The DNNs obtained better results than the domain's SoTA and a non-leaning method based on sparse approximation, with 
better restoration of the underlying solar structures, and are the new SoTA. The two training setups have different strengths and weaknesses: full supervision produces more homogeneous solar disks that are more physically plausible, and C-GAN better handles strong clouds and thick filament restoration. Thus, there is no obvious ``best setup'', and the choice of training setup depends on the intended usage.

The DNNs have been trained on fluffy cloud textures only. While they demonstrate a good generalisation to streaked cloud textures, further improvement might be obtained from training on streaked clouds in the future.
Future work may also further focus on filament restoration, and evaluate the effect of cloud removal on a subsequent image analysis \eg the automatic detection of solar structures.

\backmatter

\bmhead{Supplementary information}


Supplementary materials are provided with additional comparative results.



\bmhead{Acknowledgments}

This work was funded by ANR grant N° ANR-20-CE23-0014-01.
One of the authors (JA) thanks CNES (Centre National d'Etudes Spatiale, the French space agency) for its support in this project.

\section*{Declarations}


The data used in this study is available at \url{https://doi.org/10.5281/zenodo.7684200}. The code is available at \url{https://gitlab.lis-lab.fr/presage/cloud-removal-from-solar-imagery}.

Adeline Paiement and Jean Aboudarham contributed to the study conception and design. Data preparation was performed by Jay Paul Morgan, experiments were performed by Amal Chaoui and Jay Paul Morgan, analysis was performed by Amal Chaoui, Jay Paul Morgan and Adeline Paiement. All authors contributed to the manuscript.









\bibliography{egbib}

\end{document}


\maketitle

\section{Architecture choices for the deep learning methods}

\subsection{Number of down-/up-sampling blocks}

Tables \ref{tab:num_blocks} and \ref{tab:num_blocks_2} provide metrics for the 4 and 5 blocks configurations of both fully supervised and C-GAN networks. 

\begin{table}[h]
   \begin{center}
   \begin{tabular}{|l|cc|cc|}
       \hline
       &
      \multicolumn{2}{c|}{4 blocks} &
      \multicolumn{2}{c|}{5 blocks} \\
      & Fully sup. & C-GAN & Fully sup. & C-GAN \\
      \hline\hline
        PSNR   $\uparrow$ & 
        29.978 (0.326) & 29.430 (0.428) &
        30.472 (0.391) & 29.934 (0.437)
        \\
        SSIM   $\uparrow$ & 
        0.989 (0.001) & 0.986 (0.002) &
        0.990 (0.001) & 0.988 (0.001)
        \\
        RMSE $\downarrow$ & 
        0.040 (0.001) & 0.042 (0.002) &
        0.038 (0.001) & 0.040 (0.001)
        \\
       \hline
   \end{tabular}
   \end{center}
   \caption{Comparison of different numbers of down-/up-sampling blocks on Ca II images. Metrics are in the format mean (std) averaged over 10 runs for the fully supervised and C-GAN networks.}
   \label{tab:num_blocks}
\end{table}

\begin{table}[h]
   \begin{center}
   \begin{tabular}{|l|cc|cc|}
       \hline
       &
      \multicolumn{2}{c|}{4 blocks} &
      \multicolumn{2}{c|}{5 blocks} \\
      & Fully sup. & C-GAN & Fully sup. & C-GAN \\
      \hline\hline
        PSNR   $\uparrow$ & 
        27.401 (0.433) & 27.353 (0.203) &
        28.048 (0.815) & 27.470 (0.454)
        \\
        SSIM   $\uparrow$ & 
        0.988 (0.001) & 0.985 (0.002) &
        0.990 (0.001) & 0.988 (0.002)
        \\
        RMSE $\downarrow$ & 
        0.051 (0.001) & 0.052 (0.001) &
        0.049 (0.003) & 0.051 (0.002)
        \\
       \hline
   \end{tabular}
   \end{center}
   \caption{Comparison of different numbers of down-/up-sampling blocks on H-$\alpha$ images. Metrics are in the format mean (std) averaged over 10 runs for the fully supervised and C-GAN networks.}
   \label{tab:num_blocks_2}
\end{table}






\section{Comparative evaluation of the methods}

\begin{figure}[h]
    \centering
    \includegraphics[width=0.9\textwidth]{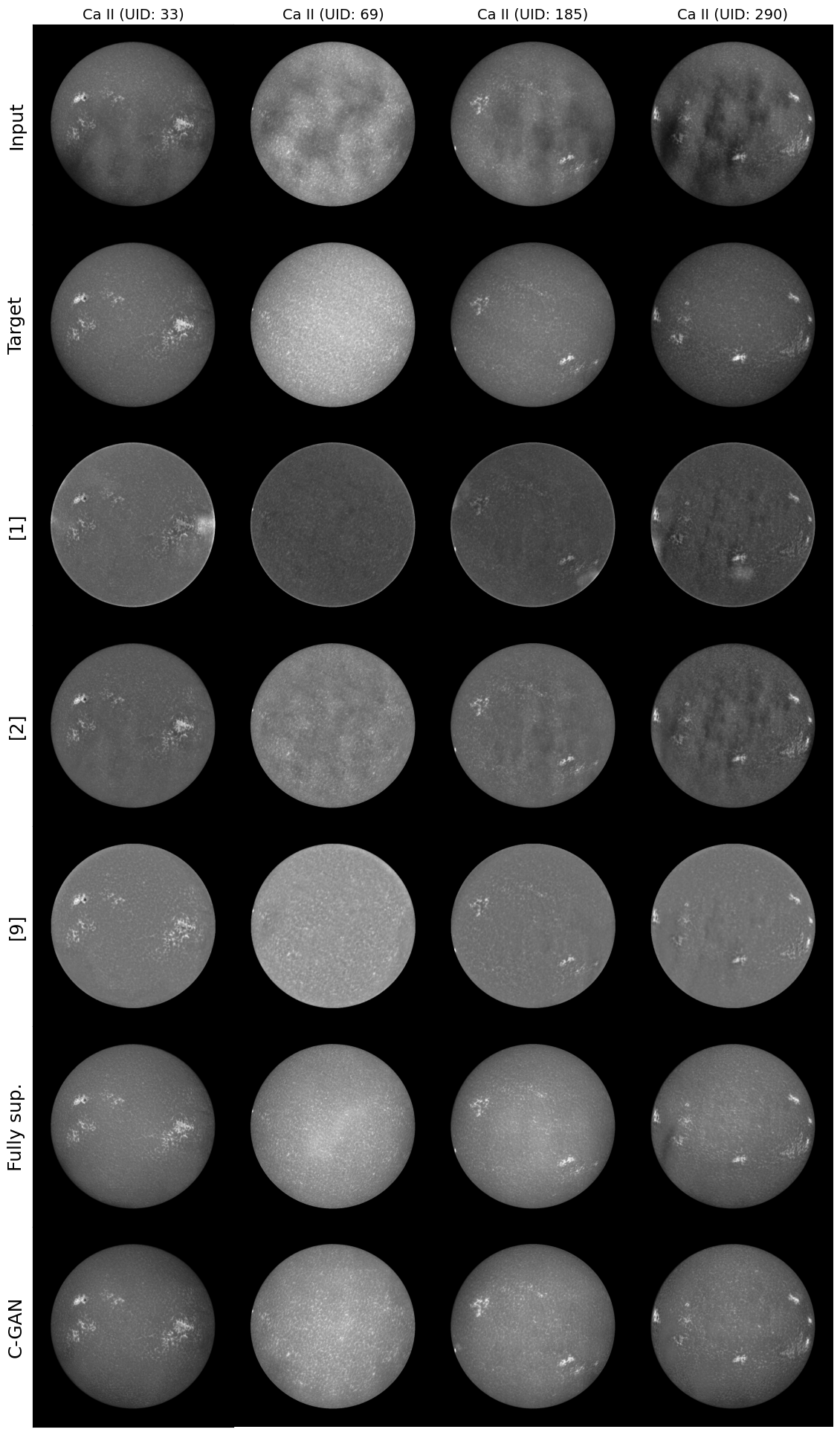}
    \caption{Example of results on synthetically generated clouds for various Ca II images.}
    \label{fig:synth_caii}
\end{figure}

\begin{figure}[h]
    \centering
    \includegraphics[width=0.9\textwidth]{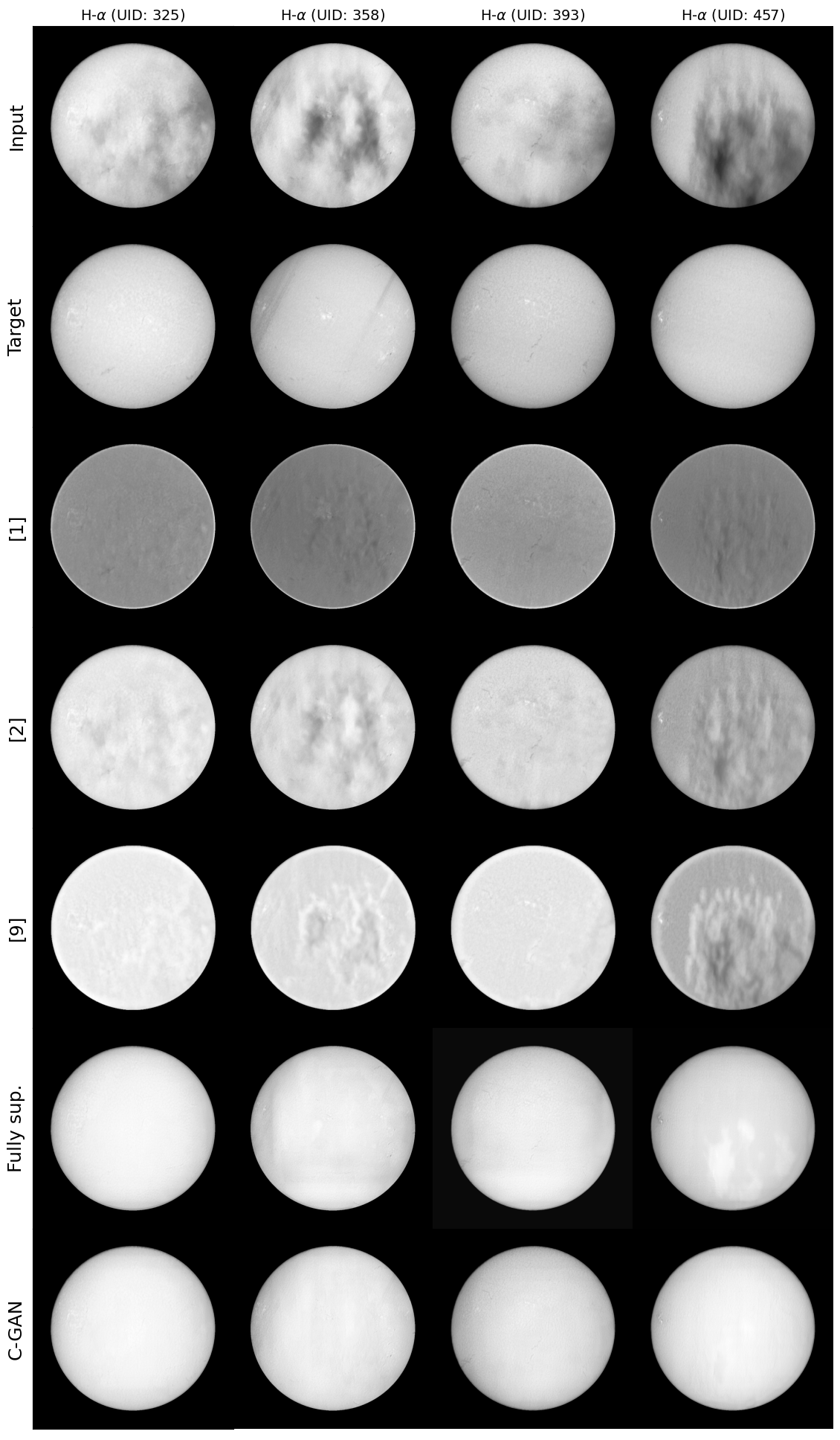}
    \caption{Example of results on synthetically generated clouds for various H-$\alpha$ images.}
    \label{fig:synth_halpha}
\end{figure}

\begin{figure}[h]
    \centering
    \includegraphics[width=\textwidth]{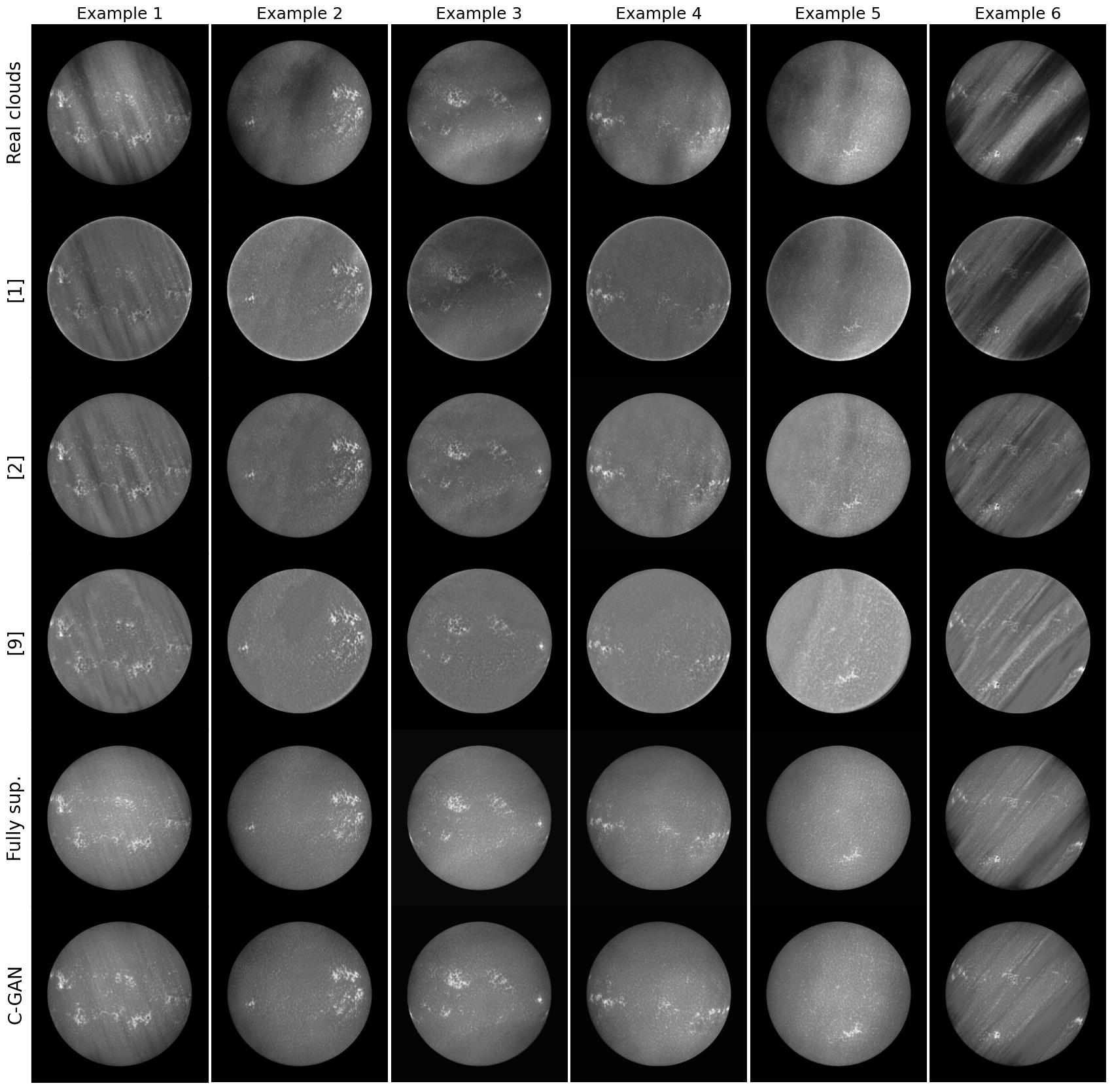}
    \caption{Example of results on Ca-II images with real clouds.}
    \label{fig:real_caii}
\end{figure}

\begin{figure}[h]
    \centering
    \begin{adjustbox}{center}
    \includegraphics[width=1.5\textwidth]{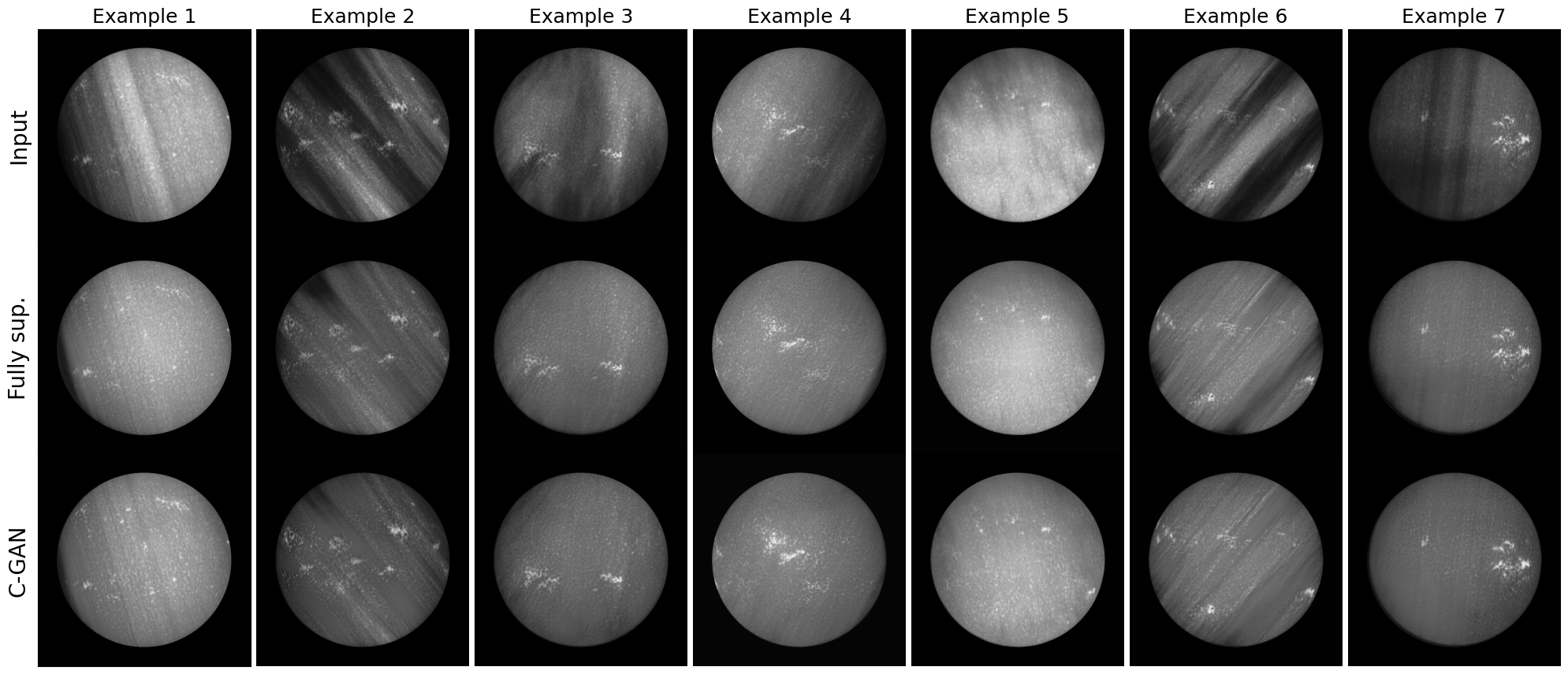}
    \end{adjustbox} 
    \caption{Further comparison of the DNNs on Ca II images with real clouds.}
    \label{fig:zoomed3}
\end{figure}

\begin{figure}[h]
    \centering
    \begin{adjustbox}{center}
    \includegraphics[width=1.2\textwidth]{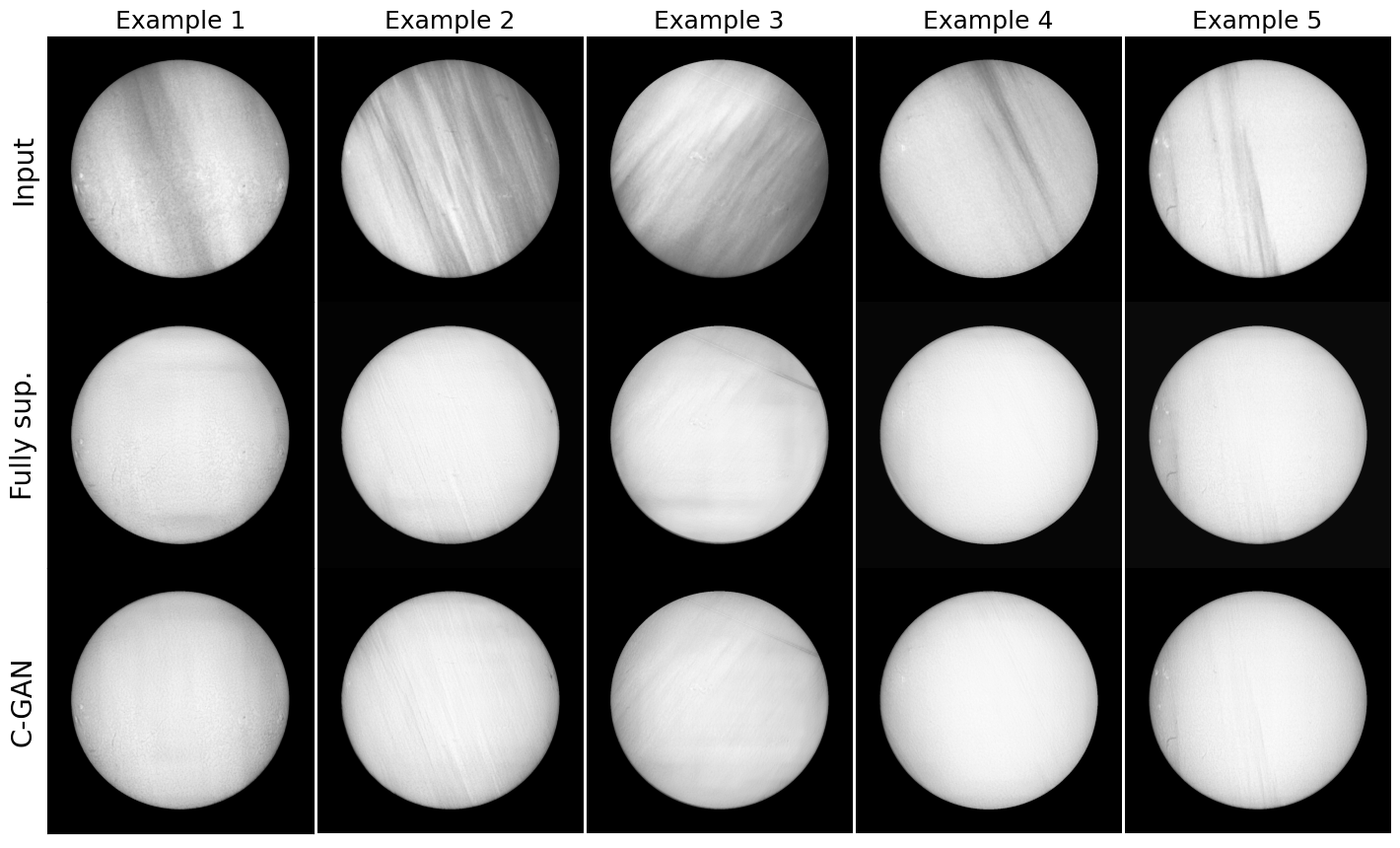}
    \end{adjustbox}
    \caption{Example of results on H-$\alpha$ images with real clouds.}
    \label{fig:zoomed2}
\end{figure}

\begin{figure}[h]
    \centering
    \includegraphics[width=0.5\textwidth]{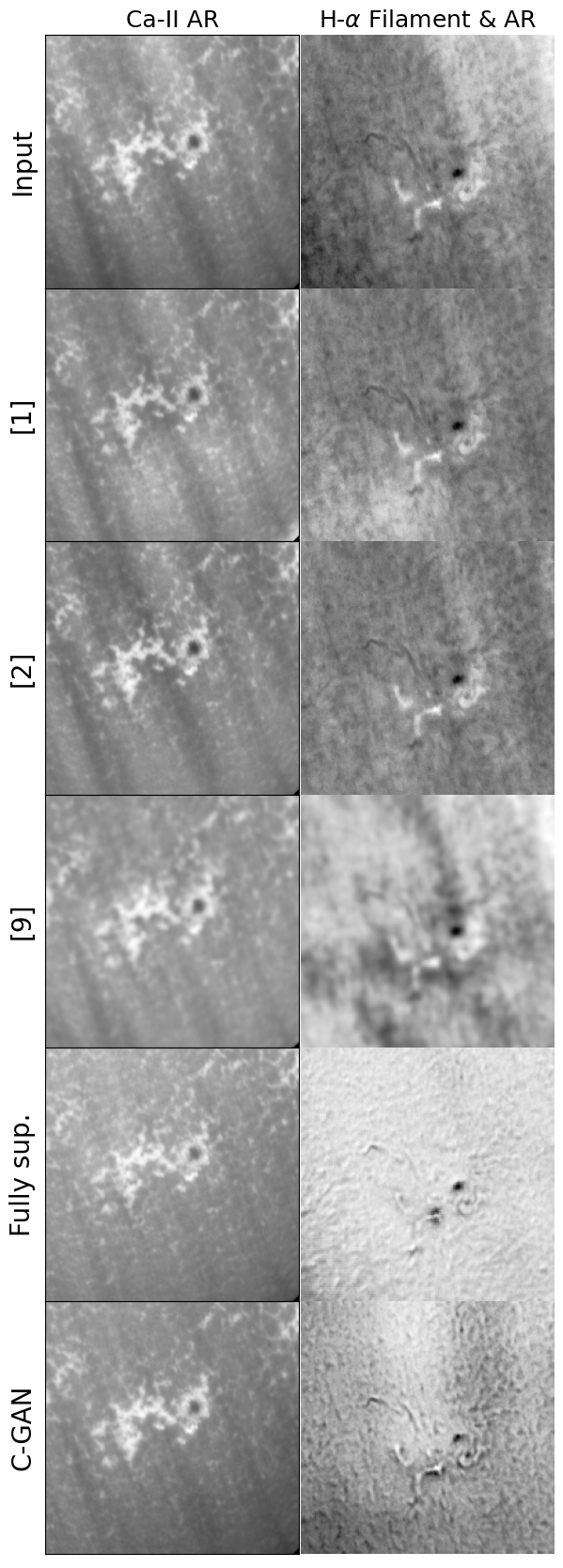}
    \caption{Example of results on zoomed regions of solar activity.}
    \label{fig:zoomed}
\end{figure}

\begin{figure}[h]
    \centering
    \begin{adjustbox}{center}
    \includegraphics[width=1.3\textwidth]{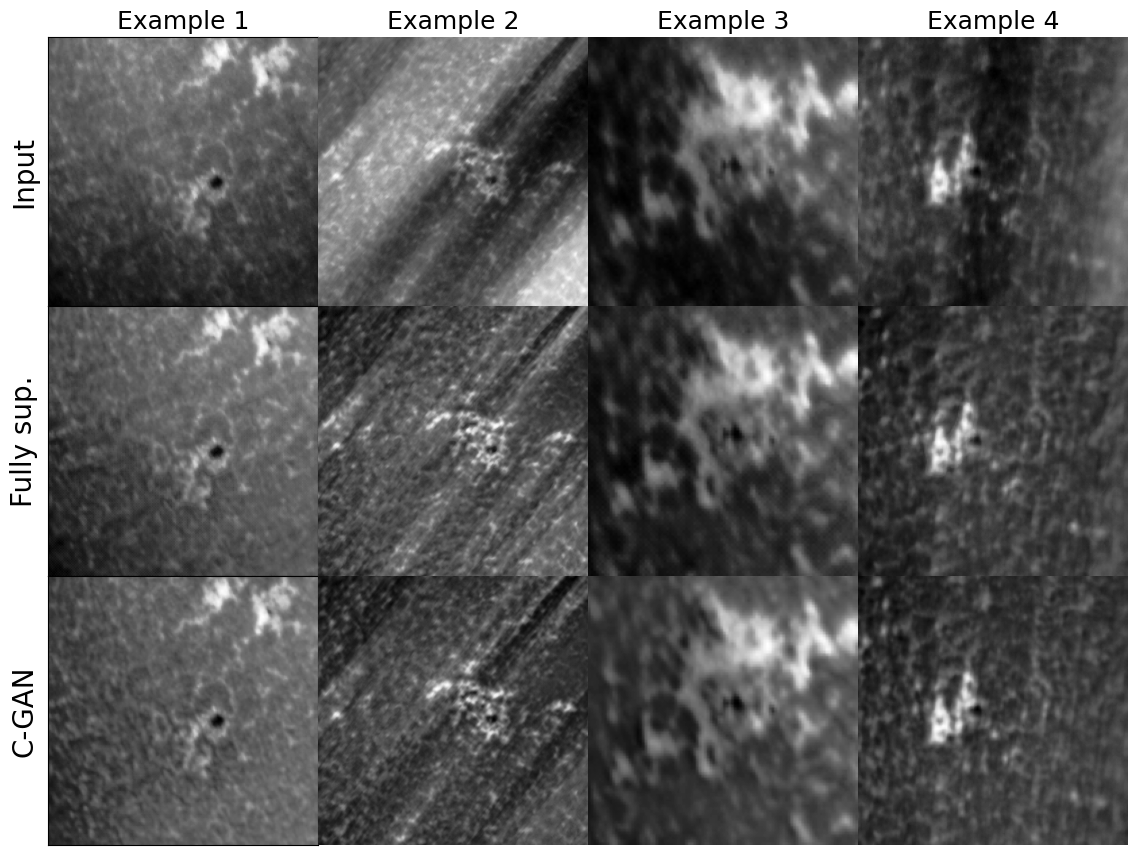}
    \end{adjustbox}
    \caption{Further comparison of the DNNs on zoomed active regions.}
    \label{fig:zoomed6}
\end{figure}

\begin{figure}[h]
    \centering
    \begin{adjustbox}{center}
    \includegraphics[width=1.3\textwidth]{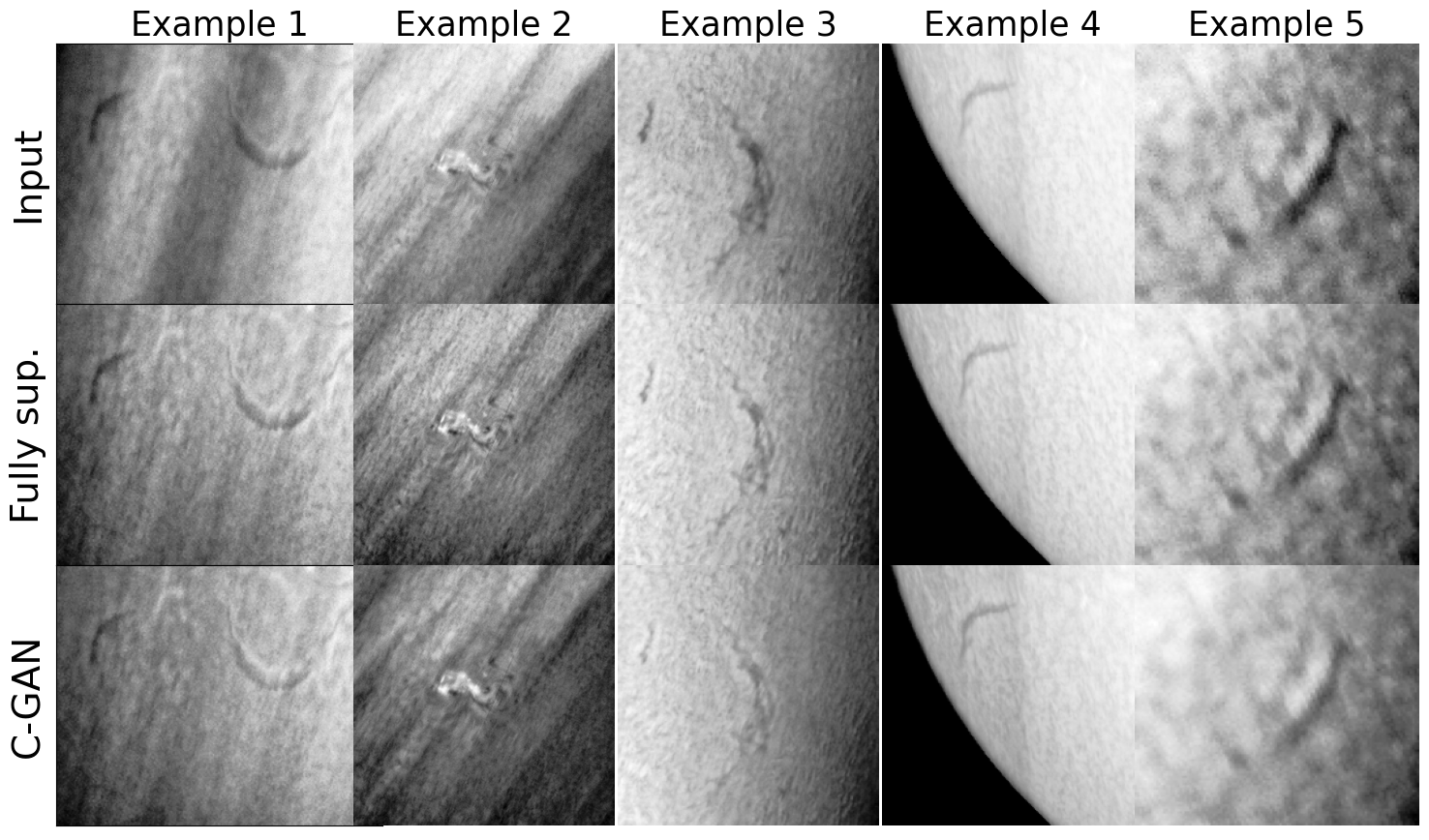}
    \end{adjustbox}
    \caption{Further comparison of the DNNs on zoomed regions of filaments.}
    \label{fig:zoomed4}
\end{figure}

Figs.~\ref{fig:synth_caii} to \ref{fig:zoomed4} present some examples of results on synthesised and real clouds, in both modalities. We display the results of a random run of each network\textcolor{red{,} since all runs produced roughly the same results.}

